\documentclass[10pt,twocolumn,letterpaper]{article}

\usepackage{cvpr}
\usepackage{times}
\usepackage{epsfig}
\usepackage{graphicx}
\usepackage{amsmath}
\usepackage{amssymb}
\usepackage{bbding}
\usepackage{tabularx}
\usepackage{multirow}
\usepackage{setspace}
\usepackage{color,subfigure}
\usepackage{mathrsfs}
\usepackage{graphics}
\usepackage{epsfig}
\usepackage{epstopdf}
\usepackage{url}
\usepackage[numbers,sort&compress]{natbib}
\usepackage{graphicx}
\usepackage{array}
\usepackage[breaklinks=true,bookmarks=false]{hyperref}

\cvprfinalcopy % *** Uncomment this line for the final submission

\def\cvprPaperID{****} % *** Enter the CVPR Paper ID here

\begin{document}

%%%%%%%%% TITLE
\title{ComDefend: An Efficient Image Compression Model to Defend Adversarial Examples}

\author{Xiaojun Jia$^1$$^,$$^2$, Xingxing Wei$^3$$^*$, Xiaochun Cao$^1$$^,$$^2$$^*$, Hassan Foroosh$^4$\\
$^1$Institute of Information Engineering, Chinese Academy of Sciences \\
$^2$Cyberspace Security Research Center, Peng Cheng Laboratory, Shenzhen 518055, China\\
$^3$Department of Computer Science and Technology, Tsinghua University,\\
$^4$Department of Electrical Engineering and Computer Science, University of Central Florida\\
{\tt\small jiaxiaojun@iie.ac.cn, xwei11@mail.tsinghua.edu.cn, caoxiaochun@iie.ac.cn, foroosh@cs.ucf.edu}}
% For a paper whose authors are all at the same institution,
% omit the following lines up until the closing ``}''.
% Additional authors and addresses can be added with ``\and'',
% just like the second author.
% To save space, use either the email address or home page, not both

\maketitle
%\thispagestyle{empty}

%%%%%%%%% ABSTRACT
\begin{abstract}
   Deep neural networks (DNNs) have been demonstrated to be vulnerable to adversarial examples. Specifically, adding imperceptible perturbations to clean images can fool the well trained deep neural networks.
   In this paper, we propose an end-to-end image compression model to defend
   adversarial examples: \textbf{ComDefend}. The proposed model consists of a compression convolutional
   neural network (ComCNN) and a reconstruction convolutional neural network (RecCNN). The
   ComCNN is used to maintain the structure information of the original image and purify
   adversarial perturbations. And the RecCNN is used to reconstruct the original image
   with high quality. In other words, ComDefend can transform the adversarial image to its clean
   version, which is then fed to the trained  classifier. Our method is a pre-processing module, and does not modify the classifier's structure during the whole process. Therefore, it can be combined with
   other model-specific defense models to jointly improve the classifier's robustness. A series of experiments conducted on MNIST, CIFAR10 and ImageNet show that the proposed method outperforms the state-of-the-art defense methods, and is
   consistently effective to protect classifiers against adversarial attacks. 
\let\thefootnote\relax\footnotetext{* Corresponding Author}
\end{abstract}

%%%%%%%%% BODY TEXT
\section{Introduction}
As we know, deep learning technique \cite{lecun2015deep}  plays the leading role in
Artificial Intelligence (AI) area, and has ushered in a new development climax
in the fields such as image recognition \cite{he2016deep},
natural language processing \cite{collobert2008unified},
and speech processing \cite{hinton2012deep}. However, Szegedy \emph{et al.} \cite{szegedy2013intriguing}
formally propose the concept of adversarial examples which bring the great danger to the neural networks. Specifically, imperceptible perturbations added to clean images
can induce networks to make incorrect predictions with high confidence during the test time,
even when the amount of perturbation is  very small, and imperceptible to human observers. What's more, \cite{kurakin2016adversarial} has proved that adversarial examples also exist
in the physical-world scenarios. The existence of adversarial examples has become a major
security concern in real-world applications of deep networks, such as self-driving cars and identity recognition, etc.

\begin{figure}
\begin{center}
   \includegraphics[width=1\linewidth]{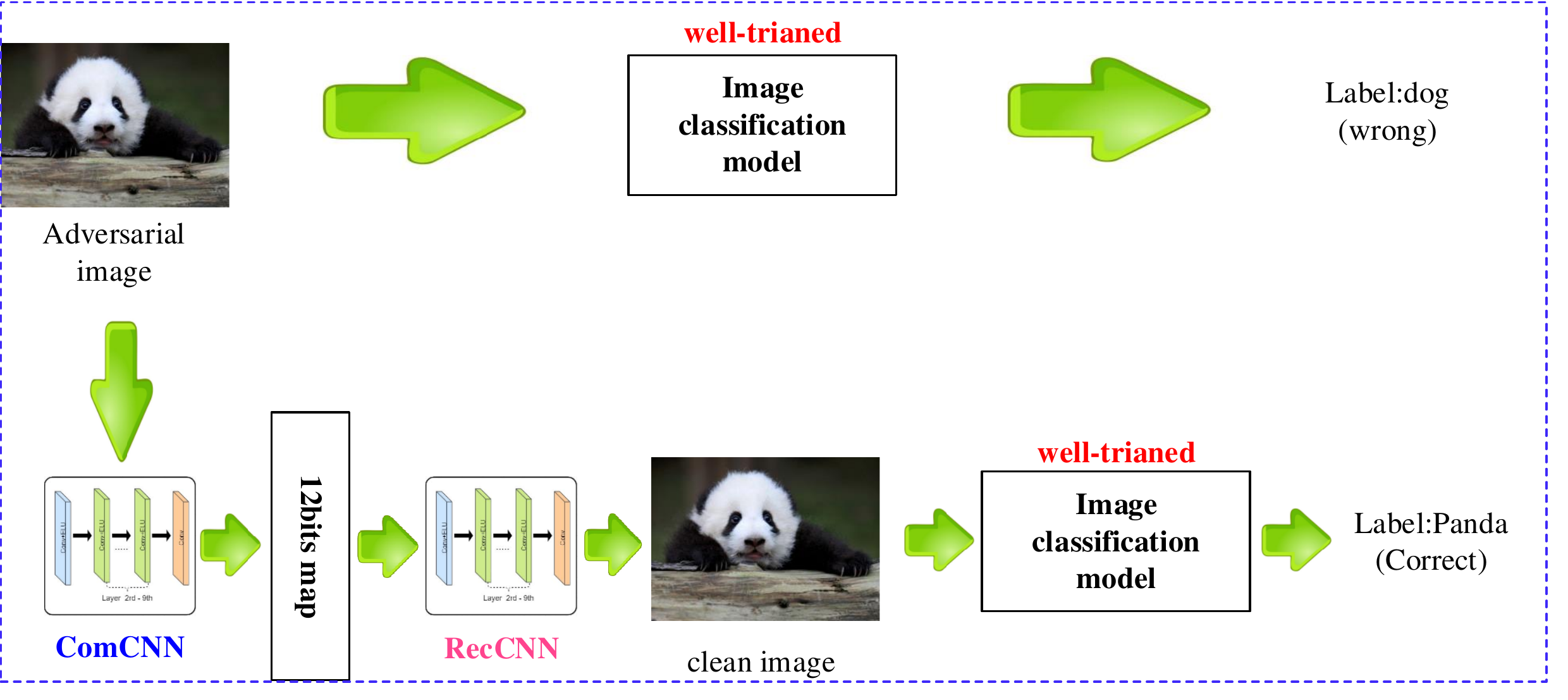}
\end{center}
   \caption{The main idea of our end-to-end image compression
   model to defend adversarial examples. The perturbation between the
   adversarial image and the original image is very tiny, but the perturbation
   is amplified during the high-level representation of the image classification model.
   We use ComCNN to remove the redundant information of the adversarial image and RecCNN to
   reconstruct the clean image. In this way,  the influence of adversarial perturbations is suppressed.
    }
\label{fig:jxj1}
\end{figure}

%-------------------------------------------------------------------------.
\begin{figure*}[t]
\begin{center}
   \includegraphics[width=1.0\linewidth]{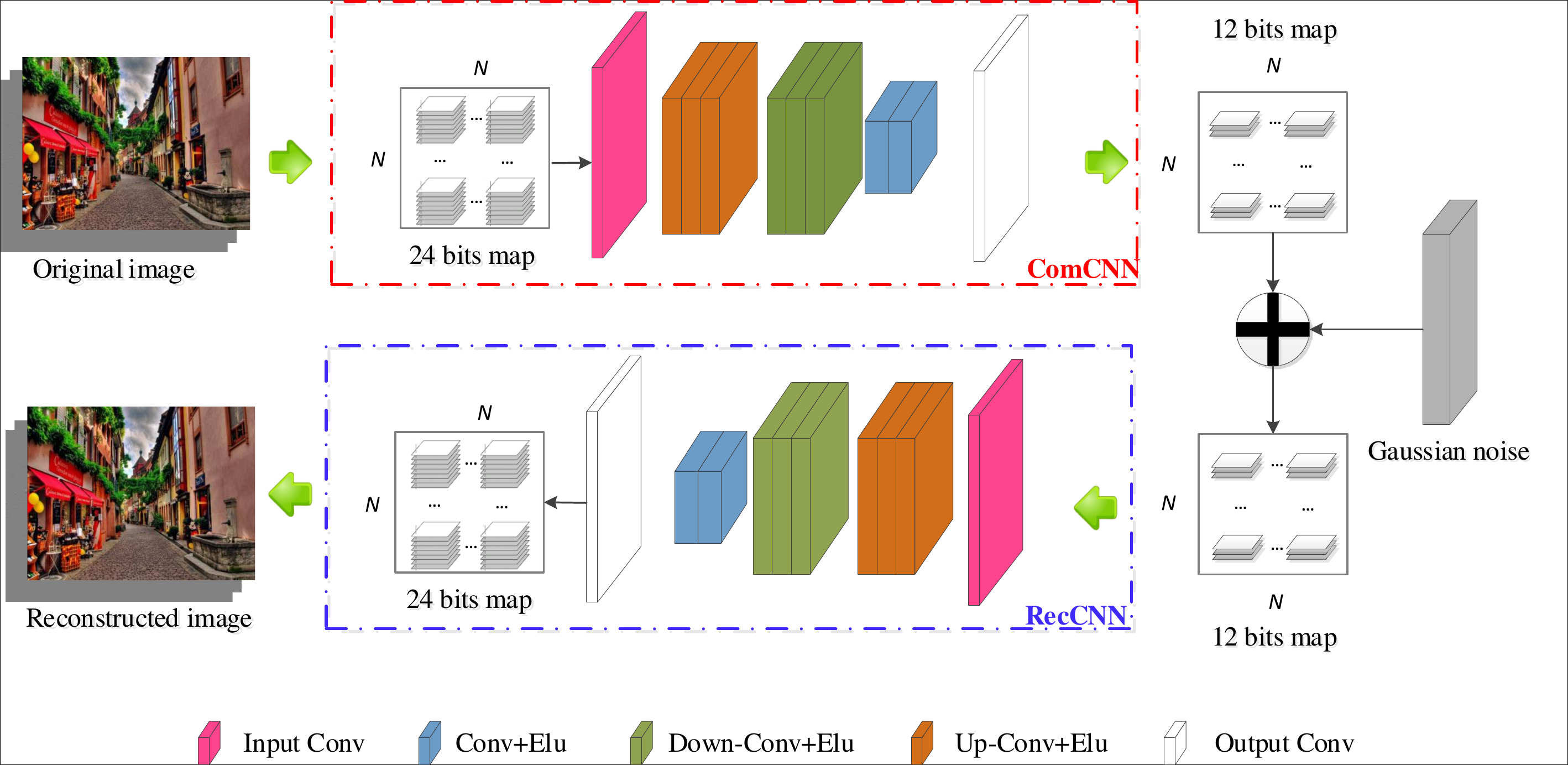}
\end{center}
   \caption{The overview of ComDefend. The ComCNN is used to preserve the main structure information of original images. The original 24 bits map for RGB three channels is compressed into 12 bits map (each channel is assigned 4 bits). And the RecCNN is  responsible for reconstructing the clean-version original images. The gaussian noise is added on the compressed compact representation to improve the reconstructed quality, and further enhance the defense ability. }
\label{fig:santan1}
\end{figure*}
%-------------------------------------------------------------------------

\par In recent years, a lot of methods defending the adversarial examples have been proposed. These methods can be roughly categorized into two classes. The first class is to enhance the robustness of neural networks itself. Adversarial training \cite{tramer2017ensemble} is a typical method among them, which injects adversarial
examples into the training data to retrain the network. Label smoothing \cite{warde2016adversarial}, which converts one-hot labels to soft targets, also belongs to this class. The second one denotes the various pre-processing methods. For example,
In \cite{song2017pixeldefend},  Song \emph{et al.} propose the PixelDefend, which can
transform the adversarial images into clean images before they are fed into the classifier. Similarly, \cite{liao2017defense}
regards the imperceptible perturbations as the noises, and designs a high-level representation guided denoiser (HGD)
 to remove these noises. HGD wins the first place in the NIPS2017 adversarial vision challenge \cite{kurakin2018adversarial}.  Generally speaking, the latter methods are more efficient because they don't need to retrain the neural networks. However, HGD still requires a lot of adversarial images when training the denoiser . Therefore, it is hard to get a good HGD in the case of few adversarial images. The main idea of PixelDefend is to simulate the distribution of image space. When the space
is too large, the result of the simulation will be bad.

%To address these issues, we propose an end-to-end image compression model which not only needs few adversarial examples but also compresses the image space. Our method is also a pre-processing step like HGD and PixelDefend.

Image compression is a low-level image transformation task. Because there is strong similarity and relevance between
neighbor pixels in the local structure, image compression can help reduce the redundant information of an image, while retaining the dominant information. Based on this observation, we devised ComDefend, which utilizes the image compression to remove adversarial perturbations or destroy the structure of adversarial perturbations. The basic idea of ComDefend is listed in Figure \ref{fig:jxj1}.

ComDefend consists of two CNN modules. The first CNN, called compression CNN (ComCNN), is used to transform the input image into a compact representation. In details, the original 24-bits pixel is compressed into 12 bits. The compact representation extracted from the input image is expected to retain the enough information of the original image. The second CNN, called reconstruction CNN (RecCNN), is used to reconstruct the original image with high quality. The ComCNN and RecCNN are finally combined into a unified end-to-end framework to learn their weights. Figure \ref{fig:santan1} gives the illustration of ComDefend. Noted that ComDefend is trained on the clean images. In this way, the network will learn the distribution of clean images, and thus can reconstruct a clean-version image from the adversarial image. Compared with HGD and PixelDefend, ComDefend doesn't require the adversarial examples in training phase, and thus reduces the computation cost. In addition, ComDefend is performed on an image with the patch-by-patch manner instead of the whole image, which improve the processing efficiency. The code is released at https://github.com/jiaxiaojunQAQ/Comdefend.git.

\par In summary, this paper has the following contributions:
\par 1) We propose the \emph{ComDefend}, an end-to-end image compression model to defend adversarial examples. The ComCNN extracts the structure information of the original image and removes the imperceptible perturbations. The RecCNN reconstructs the input image with high quality. During the whole process, the deployed model is not modified.
\par 2) We design a unified learning algorithm to simultaneously learn the weights of two CNN modules within \emph{ComDefend}. In addition, we find that adding gaussian noise to the compact representation can help reconstruct better images, and further improve the defending performance.
\par 3) Our method greatly improves the resilience across a wide variety of strong attacking methods, and defeats the state-of-the-art defense models including the winner of NIPS 2017 adversarial challenge. 

\par The remainder of this paper is organized as follows. Section 2 briefly reviews the related work. Section 3 introduces the details of the proposed ComDefend. Section 4 shows a series of experimental results. Finally, Section 5 shows the conclusion.

%------------------------------------------------------------------------
\section{Related work}
\par We investigate the related work from two aspects:
 Attack methods to generate adversarial examples, and Defensive methods to resist adversarial examples.

\subsection{Attack methods}

\par In \cite{goodfellow6572explaining}, Goodfellow \emph{et al}. propose the Fast Gradient Sign Method (FGSM). An adversarial example is produced by adding increments in the gradient direction of the loss gradient. After that, Basic Iterative Method (BIM) which is the improved version of the FGSM, is proposed in \cite{kurakin2016adversarial}.
Compared with FGSM, BIM performs multiple steps. This method is also called Projected
Gradient Descent (PGD) in \cite{madry2017towards}. To deal with the selection
of parameters in FGSM, in \cite{moosavi2016deepfool}, Moosavi-Dezfooli
\emph{et al}. propose to use an iterative linearization of the classifier and geometric formulas to generate an adversarial example. In \cite{carlini2017towards},
Carlini-Wagner \emph{et al}. design an efficient optimization objective (C\&W) to find the smallest perturbations. The C\&W  can reliably produce samples correctly classified by human subjects but misclassified in specific targets by the well-trained classifier.

%-------------------------------------------------------------------------

%-------------------------------------------------------------------------
\subsection{Defensive methods}

\par In \cite{tramer2017ensemble}, Adversarial training adds the adversarial images generated
by different attack methods to the training image dataset. The growth of the training image dataset
makes the image classification model easier to simulate the distribution of the entire image space.
And in \cite{warde201611}, Warde-Farley and Goodfellow propose label smoothing method which
uses soft targets to replace one-hot labels. The image classifier is trained on the one-hot labels at first,
and then the soft targets are generated by the well-trained image classifier. In \cite{xu2017feature},
Xu \emph{et al}. propose to use feature squeezing methods which include the color bit depth of each pixel and
spatial smoothing to achieve defend adversarial examples. PiexlDefend is proposed in \cite{song2017pixeldefend}.
The basic idea of PiexlDefend is to purify input images before they are fed to the image classifier.
In \cite{liao2017defense}, the authors propose a high-level representation guided
denoiser(HGD) method to defend adversarial examples. The proposed model is trained on the training
dataset which includes 210k clean and adversarial images.

%-------------------------------------------------------------------------
\section{End-to-end image compression model}

\subsection{The basic idea of ComDefend}
Let us first look back at the reason of adversarial examples. The adversarial examples are generated by adding some imperceptible perturbation to the clean images.
The added perturbation is too slight to be perceptible to humans. However, when the adversarial examples are fed to a deep learning network, the effect of the imperceptible perturbation increases rapidly along with the deepth of the network. Therefore, the carefully designed perturbation will fool powerful CNNs. More specifically, from previously related researches, we can regard the imperceptible perturbation as the noise
with the particular structure. Kurakin \emph{et al}. in \cite{kurakin2016adversarial}
consider that this kind of noise which can fool powerful CNNs exists in the real
world. In other words, the perturbations do not affect the structure information of the original image. The imperceptible perturbations can be considered as the redundant information of the images. From this point of view, we can use the characteristics of image redundancy information in image compression model to  defend adversarial examples.

\par In order to remove the imperceptible perturbations or break up the particular structure of the imperceptible perturbations, we propose an end-to-end image compression model which not only compresses the input image but also transforms the input image to a clean image. As shown in Figure \ref{fig:santan1}, the image compression model contains the compression and reconstruction processes. During the compression process, the ComCNN extracts the image structure information and removes the redundant information of the image. During the reconstruction process, the RecCNN reconstructs the input image without the adversarial perturbations.
In particular, the ComCNN compress the 24-bits pixel image into 12 bits. That is to say, the 12-bits pixel image removes the redundancy information of the original image
and preserves the main information of the original image. And thus the RecCNN use the 12-bits pixel image to reconstruct the original image. During the whole process, we hope that the 12-bits pixel images extracted from the original image and adversarial example are as same as possible. Therefore, we can transform the adversarial example into the clean image.

%-------------------------------------------------------------------------

\begin{figure*}
\begin{center}
   \includegraphics[width=1.0\linewidth]{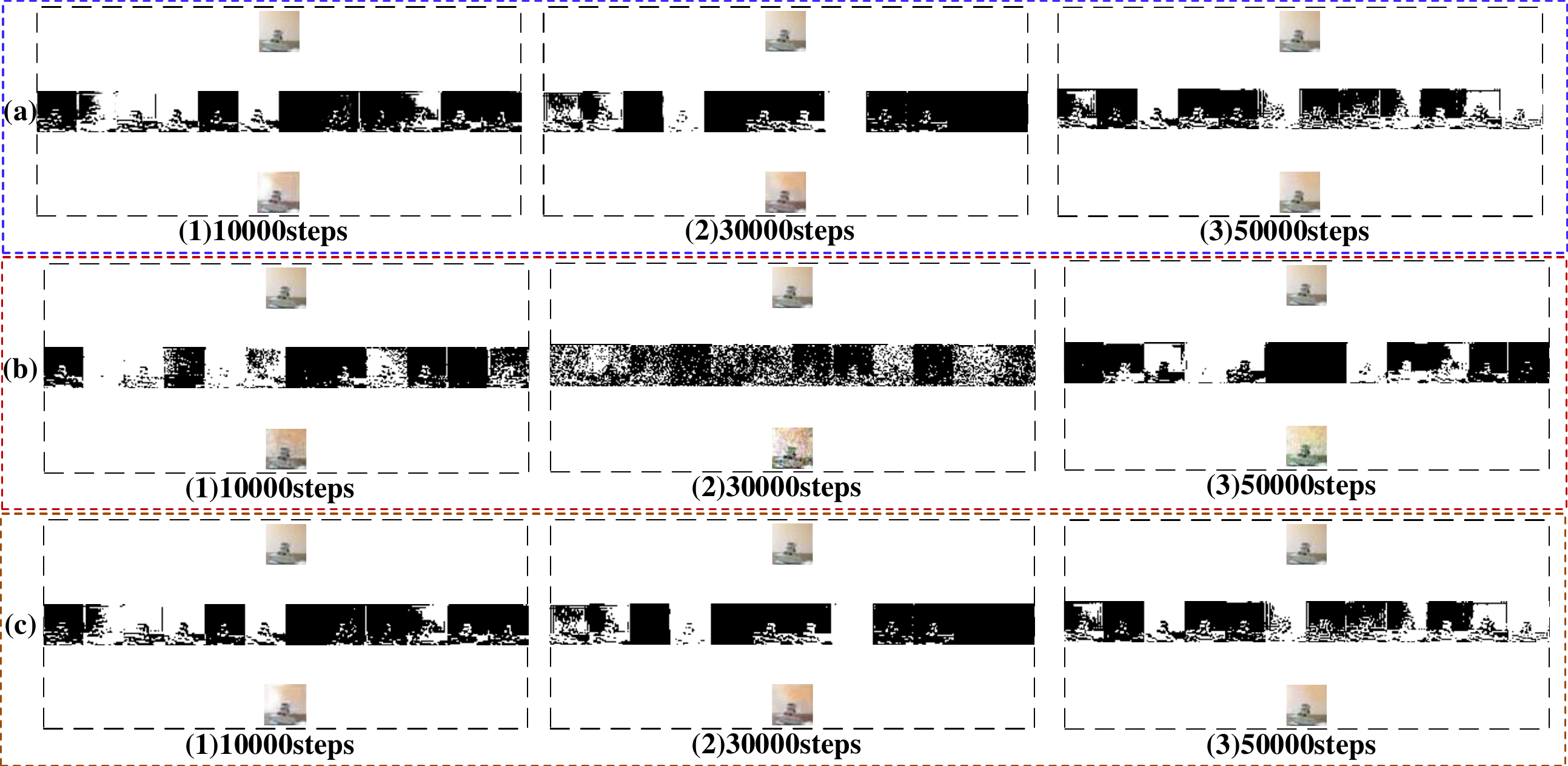}
\end{center}
   \caption{The comparison results in ComDefend whether to add gaussian noises. In each subfigure, The top images are original images, The middle images are the compressed 12bits maps and the bottom images are the reconstructed images. (a) ComDefend reconstructs the image through the un-binarized 12bits map. (b) Without gaussian noises, ComDefend reconstructs the image through the binarized 12bits map. (c) With gaussian noises, ComDefend reconstructs the image through the binarized 12bits map.  We see the reconstructed quality in (c) is the same with that in (a). That means the increment information of un-binarized maps are actually noises. Therefore, when gaussian noises are added on the binarized maps, the better images are reconstructed. }
\label{fig:santan2}
\end{figure*}

\begin{table}[t]
\centering
\footnotesize{
\caption{  Hyperparameters of the ComCNN Layers }
 \label{table:crossstreetmAPresult0}
\begin{tabular}{| c c c c c | }
  \hline
  layer &  type & output channels & input channels & filter size  \\
 \hline
    1st layer  & conv+ELU  & 16 & 3 & \begin{math} 3\times 3\end{math} \\
    2nd layer  & Conv+ELU  & 32 & 16 & \begin{math} 3\times 3\end{math} \\
    3rd layer  & Conv+ELU  & 64 & 32 & \begin{math} 3\times 3\end{math} \\
     4th layer  & Conv+ELU  & 128 & 64 & \begin{math} 3\times 3\end{math} \\
     5th layer  & Conv+ELU  & 356 & 128 & \begin{math} 3\times 3\end{math} \\
     6th layer  & Conv+ELU  & 128 & 256 & \begin{math} 3\times 3\end{math} \\
     7th layer  & Conv+ELU  & 64 & 128 & \begin{math} 3\times 3\end{math} \\
     8th layer  & Conv+ELU  & 32 & 64 & \begin{math} 3\times 3\end{math} \\
     9th layer  & Conv & 12 & 32 & \begin{math} 3\times 3\end{math} \\
    \hline
\end{tabular}
}
\end{table}

\begin{table}[t]
\centering
\footnotesize{
\caption{  Hyperparameters of the RecCNN Layers }
 \label{table:crossstreetmAPresult1}
\begin{tabular}{| c c c c c | }
  \hline
  layer &  type & output channels & input channels & filter size  \\
 \hline
    1st layer  & Conv+ELU  & 32 & 12 & \begin{math} 3\times 3\end{math} \\
    2nd layer  & Conv+ELU  & 64 & 32 & \begin{math} 3\times 3\end{math} \\
    3rd layer  & Conv+ELU  & 128 & 64 & \begin{math} 3\times 3\end{math} \\
     4th layer  & Conv+ELU  & 256 & 128 & \begin{math} 3\times 3\end{math} \\
     5th layer  & Conv+ELU  & 128 & 256 & \begin{math} 3\times 3\end{math} \\
     6th layer  & Conv+ELU  & 64 & 128 & \begin{math} 3\times 3\end{math} \\
     7th layer  & Conv+ELU  & 32 & 64 & \begin{math} 3\times 3\end{math} \\
     8th layer  & Conv+ELU  & 16 & 32 & \begin{math} 3\times 3\end{math} \\
     9th layer  & Conv & 3 & 16 & \begin{math} 3\times 3\end{math} \\
    \hline
\end{tabular}
}
\end{table}

\subsection{Structure of the ComCNN}
ComCNN consists of 9 weight layers, which can compress the input image into the 12-bits pixel image. That is
to say, the main structure information of the input image is reserved and the redundancy information
including the imperceptible perturbation of the input image is removed. The combination
of convolution and ELU \cite{clevert2015fast} are used in ComCNN. As shown in Table \ref{table:crossstreetmAPresult0},
ComCNN consists of two components, the first one is used to extract the features of the
original image and generate 256 feature maps. The 1st to the 4th layers which
consist of 32 filters of size \begin{math}3 \times 3 \times 3 \end{math}, 64 filters of
size \begin{math}3 \times 3 \times 32\end{math}, 128 filters  of size \begin{math}3 \times 3 \times 64\end{math}
and 256 filters of size \begin{math}3 \times 3 \times 128\end{math} are the main part of the
first component. And the ELU nonlinearity is used as an activation function. The second one is
used to downscale and enhance the features of the input image. The 5th to the 9th layers which
consist of 128 filters of size \begin{math}3 \times 3 \times 256\end{math}, 64 filters of size
\begin{math}3 \times 3 \times 128\end{math}, 64 filters of size \begin{math}3 \times 3 \times 64\end{math},
32 filters of size \begin{math}3 \times 3 \times 64\end{math} and 3 filters of
size \begin{math}3 \times 3 \times 32\end{math} are the main part of the second component. The
ComCNN is used to extract the features of the original image and construct the compact representation.
%-------------------------------------------------------------------------

%-------------------------------------------------------------------------
\subsection{Structure of the RecCNN}
RecCNN consists of 9 weight layers, which is used to reconstruct the original image without
the imperceptible perturbation. As shown in Table \ref{table:crossstreetmAPresult1}, For the 1st layer
to the 9th layers, 32 filters of size \begin{math}3 \times 3 \times 12\end{math},
64 filters of size \begin{math}3 \times 3 \times 32\end{math}, 128 filters of size
\begin{math}3 \times 3 \times 64\end{math}, 256 filters of size
\begin{math}3 \times 3 \times 128\end{math}, 128 filters of size
\begin{math}3 \times 3 \times 256\end{math}, 64 filters of size
\begin{math}3 \times 3 \times 128\end{math}, 64 filters of size
\begin{math}3 \times 3 \times 64 \end{math}, 32 filters of size
\begin{math}3 \times 3 \times 64 \end{math} and 3 filters of size
\begin{math}3 \times 3 \times 32 \end{math} are used, and ELU is added. The
RecCNN makes use of the compact representation to reconstruct the output image.
The output image has the fewer perturbations than the input image. That is to say,
the output image can break up the particular structure of the perturbations. In this way, the compression model can defend the adversarial examples.
%-------------------------------------------------------------------------

%-------------------------------------------------------------------------

%-------------------------------------------------------------------------
%-------------------------------------------------------------------------
\subsection{Loss functions}
As for ComCNN, the goal of the ComCNN is to use more 0 to encode the image information. Therefore,
the loss function of the ComCNN can be defined as:
\begin{equation}\label{mvr}
L_{1}(\theta_{1})= \lambda\|Com(\theta_{1},x)\|^{2},
\end{equation}
where \begin{math}\theta_{1}\end{math} is the trainable parameter of the ComCNN,
$Com()$ represents the ComCNN,
and \begin{math}\lambda\end{math} is a super parameter which we use a large number experiments to certify.
Please refer to Section 4 for more details.
\par As for RecCNN, the goal of the RecCNN is expected to reconstruct the original image with high quality.
Therefore, we use MSE to define the loss function of the RecCNN:
\begin{equation}\label{mvr}
L_{2}(\theta_{2})= \frac{1}{2N}\Sigma\|Rec(\theta_{2},Com(\theta_{1},x)+\varphi)-x\|^{2},
\end{equation}
where \begin{math}Com()\end{math} is the ComCNN, \begin{math}\theta_{2}\end{math} represents the trainable parameter of the RecCNN,
 \begin{math}Rec()\end{math} represents the RecCNN, \begin{math}\theta_{1}\end{math} represents the trained parameter of the ComCNN.
 And \begin{math} \varphi \end{math} represents the random Gaussian noise.
\par In order to make the compression model more effective, we design a unified loss function to
simultaneously update the parameters of ComCNN and RecCNN. It is defined as:
\begin{equation}\label{mvr}
\begin{split}
L(\theta_{1},\theta_{2})= \frac{1}{2N}\Sigma\|Rec(\theta_{2},Com(\theta_{1},x)+\varphi)-x\|^{2}\\
+\lambda\|Com(\theta_{1},x)\|^{2}.
\end{split}
\end{equation}
\par According to this loss function, it is clear that both ComCNN and RecCNN work together to
resist the noise attack. The parameters \begin{math}\theta_{1},\theta_{2} \end{math} are upgraded
at the same time during the model training.

%-------------------------------------------------------------------------
\subsection{Learning algorithm}
In order to train the compression model, we design a unified learning algorithm for
both ComCNN and RecCNN. The optimization goal for ComDefend is formulated as:
\begin{equation}\label{mvr}
\begin{split}
(\theta_{1},\theta_{2})=arg \min( \frac{1}{2N}\Sigma\|Rec(\theta_{2},Rec(\theta_{1},x)+\varphi)-x\|^{2}\\
+\|Com(\theta_{1},x)\|^{2}),
\end{split}
\end{equation}
where \begin{math}x\end{math} is the input image. \begin{math}\varphi\end{math} represents
the random Gaussian noise. \begin{math}\theta_{1}\end{math} and \begin{math}\theta_{2}\end{math}
are the parameters of ComCNN and RecCNN respectively. \begin{math}Com()\end{math} represents the
ComCNN and \begin{math}Rec()\end{math} represents RecCNN.
\par During the whole process, the ComCNN encodes the input image \begin{math}x\end{math} into
a same size image \begin{math}y\end{math} with each pixel occupies 12 floats. Then the sigmoid
function is used to limit the image \begin{math}y\end{math} to between 0 and 1. Note that, the sigmoid output makes use of the different shades of gray information to represent the input image instead of 0 and 1. And RecCNN can reconstruct the original image through these shades of gray
information. If these shades of gray information are binarized, the main structure information of
original image is completely lost. In order to deal with this problem, we propose to use the noise attack.
\par In particular, we add the random Gaussian noise \begin{math}\varphi\end{math} (the mean of the gaussian noise is 0 and the variance of the gaussian noise is \begin{math}\varphi\end{math}) to the output before
the sigmoid function. The information encoded with 0 and 1 is easier to resist the noise attack. Therefore,
during the training, the compression model learns to use the binary information to defend the noise attack.
As shown in Figure \ref{fig:santan2}, we can see that adding the random gaussian noise contributes to improving the performance of the compression model.
In addition, We choose the compression bits mainly according to the reconstructed performance. We try different compression bits in Table \ref{table:crossstreetmAPresult8}, and find the 12 bits show the best PSNR reconstructed performance.

\begin{table}
\centering
\footnotesize{
\caption{ The experiments versus selection of compression bits}
 \label{table:crossstreetmAPresult8}
\resizebox{230pt}{12pt}{
\begin{tabular}{|c|c|c|c|c|c|}
\hline
Compressed bits & 8     & 10    & 12    & 14    & 16    \\ \hline
PSNR           & 31.01 & 31.01 & \textbf{31.78} & 28.77 & 30.95 \\ \hline
\end{tabular}
}}
\end{table}
%-------------------------------------------------------------------------
\subsection{Network implementation}
The weights of the ComCNN and the RecCNN are initialized by using the method
in \cite{he2015delving}. We also use Adam algorithm \cite{kingma2014adam} with parameters setting
\begin{math} \alpha = 0.001,  \beta_{1} = 0.9,  \beta_{2} = 0.999 \end{math}
and \begin{math} \varepsilon = 10^{-8} \end{math} to upgrade the weights of
the compression model. After the hyperparameters \begin{math} \gamma \end{math}
and \begin{math} \lambda \end{math} being confirmed, we train ComCNN and RecCNN for 30 epochs using a batch size of 50. The learning rate is decayed exponentially
from 0.01 to 0.0001 for 30 epochs.

%-------------------------------------------------------------------------
\begin{table*}[t]
\centering
\footnotesize{
\caption{ THE SELECTION OF PARAMETERS IN THE OUR PROPOSED METHOD . }
 \label{table:crossstreetmAPresult3}
\begin{tabular}{| c | c | c | c | c | c | c |c | c | c | c |}
  \hline
   & \begin{math} \varphi = 1.0  \end{math}  & \begin{math} \varphi = 10.0 \end{math} &  \begin{math} \varphi = 20.0 \end{math} & \begin{math} \varphi = 25.0 \end{math} & \begin{math} \varphi = 30.0 \end{math} & \begin{math}  \varphi = 35.0 \end{math} & \begin{math} \varphi = 40.0 \end{math} & \begin{math} \varphi = 50.0 \end{math} & \textbf{Average}\\
 \hline

    \begin{math} \lambda = 0.01 \end{math} & 68.39\% &90.22\% & 86.69\% & 87.52\% & 86.12\% & 85.42\% & 86.22\% & 86.71\% & 84.66\% \\
    \hline
    \begin{math} \lambda = 0.001 \end{math}  & 88.99\% & 89.64\% & 72.23\% & 90.23\% & 91.09\% & 90.41\% & 90.55\% & 90.56\% & 87.96\% \\
    \hline
    \begin{math} \lambda = 0.0001 \end{math} &  89.61\%& 90.77\% &  \textbf{91.82\%} & 90.98\% & 89.45\% & 91.24\% & 90.61\% & 90.33\% & 90.60\% \\
    \hline
    \begin{math} \lambda = 0.00001 \end{math} &  89.06\% & 91.65\% &  90.99\% & 91.37\% & 90.74\% & 91.05\% & 90.25\% & 90.65\% & 90.72\% \\
    \hline
    \begin{math} \lambda = 0.0 \end{math} &  90.00\% & 90.39\% &  91.45\% &91.27\% & 91.01\% & 90.88\% & 88.18\% & 90.10\% & 90.41\% \\
    \hline
     \textbf{Average} &  85.19\% & 90.53\% &  86.63\% & 90.27\% & 89.88\% & 89.80\% & 89.16\% & 89.67\% & 88.89\% \\
\hline
\end{tabular}
}
\end{table*}

%-------------------------------------------------------------------------
\section{Experimental results and analysis}
In this section, in order to evaluate the performance of the
proposed method, we conduct several experiments, which
include: generation of adversarial examples, selection of hyper parameters in neural networks, image classification with the proposed method, comparisons with other defensive methods and performance analysis. The proposed method can significantly perform well against the state-of-the-art adversarial attacks.
%-------------------------------------------------------------------------
\subsection{Datasets for training and testing}
In order to clearly verify our proposed method, the ComCNN and RecCNN training are based on the 50,000 clean (not perturbed) images of the CIFAR-10 dataset \cite{krizhevsky2010cifar}. For
testing, we use 10,000 testing images in the CIFAR-10 dataset,
10,000 testing images in the Fashion-mnist \cite{xiao2017fashion} and 1000 random images of the imagenet dataset \cite{deng2009imagenet}. We also train ResNet \cite{he2016deep} which is one of the state-of-the-art deep neural network image classifiers in recent years on these three datasets.
%-------------------------------------------------------------------------

%-------------------------------------------------------------------------
\subsection{Adversarial examples}
In the literature, three common distance metrics are used for generating adversarial examples: \begin{math}L_{0}, L_{2}, L_{\infty}\end{math}.
\begin{math}L_{0}\end{math} represents the number of the different pixels
between the clean image and adversarial example. \begin{math}L_{2}\end{math}
measures the standard Euclidean distance between the clean
image and adversarial example. \begin{math}L_{\infty}\end{math} represents
the maximum value of the imperceptible perturbation in the adversarial example.
In \cite{warde2016adversarial}, Goodfellow et al. argue to use \begin{math}L_{\infty}\end{math} to
construct the adversarial examples. And the related research literature main
use \begin{math}L_{2}\end{math} and \begin{math}L_{\infty}\end{math} to conduct
related researches. Therefore, we make use of \begin{math}L_{2}\end{math}
and \begin{math}L_{\infty}\end{math} to achieve the adversarial attacks.
In particular, we use the \begin{math}L_{\infty}\end{math} distance metric to
achieve FSGM, BIM and DeepFool adversarial attacks and the \begin{math}L_{2}\end{math}
distance metric to achieve C\&W adversarial attacks.

%-------------------------------------------------------------------------

\begin{table*}[t]
\footnotesize{
\caption{ THE RESULT OF COMPARISONS WITH OTHER DEFENSIVE METHODS(CIFAR-10 ,$L_{\infty}$ = 2/8/16) }
 \label{table:crossstreetmAPresult4}
\resizebox{500pt}{45pt}{

\begin{tabular}{|c|c|c|c|c|c|c|c|}
\hline
\multirow{2}{*}{Network}  & \multicolumn{2}{c|}{\multirow{2}{*}{Defensive method}}   & \multirow{2}{*}{Clean} & \multirow{2}{*}{FGSM} & \multirow{2}{*}{BIM} & \multirow{2}{*}{DeepFool} & \multirow{2}{*}{C\&W} \\
                          & \multicolumn{2}{c|}{}                                 &                        &                       &                      &                           &                     \\ \hline
\multirow{8}{*}{Resnet50} & \multirow{5}{*}{In training time} & Normal            & 92\%/92\%/92\%         & 39\%/20\%/18\%        & 08\%/00\%/00\%       & 21\%/01\%/01\%            & 17\%/00\%/00\%      \\ \cline{3-8}
                          &                                   & Adversarial FGSM  & 91\%/91\%/91\%         & 88\%/91\%/91\%        & 24\%/07\%/00\%       & 45\%/00\%/00\%            & 20\%/00\%/07\%      \\ \cline{3-8}
                          &                                   & Adversarial BIM   & 87\%/87\%/87\%         & 80\%/52\%/34\%        & 74\%/32\%/06\%       & 79\%/48\%/25\%            & 76\%/42\%/08\%      \\ \cline{3-8}
                          &                                   & Label Smoothing   & 92\%/92\%/92\%         & 73\%/54\%/28\%        & 59\%/08\%/01\%       & 56\%/20\%/10\%            & 30\%/02\%/02\%      \\ \cline{3-8}
                          &                                   & \textbf{Proposed method}   & \textbf{92\%/92\%/92\%}         & \textbf{89\%/89\%/87\%}        & \textbf{84\%/47\%/40\%}       & \textbf{90\%/90\%/90\%}            & \textbf{91\%/90\%/90\%}      \\ \cline{2-8}
                          & \multirow{3}{*}{In test time}     & Feature Squeezing & 84\%/84\%/84\%         & 31\%/20\%/18\%        & 13\%/00\%/00\%       & 75\%/75\%/75\%            & 78\%/78\%/78\%      \\ \cline{3-8}
                          &                                   & PiexlDefend       & 85\%/85\%/88\%         & 73\%/46\%/24\%        & 71\%/\textbf{46\%}/25\%       & 80\%/80\%/80\%            & 78\%/78\%/78\%      \\ \cline{3-8}
                          &                                   & \textbf{Proposed method}   & \textbf{91\%/91\%/91\%}         & \textbf{86\%/84\%/83\%}        & \textbf{78\%}/41\%/\textbf{34\%}       & \textbf{88\%/88\%/88\%}            & \textbf{89\%/87\%/87\%}      \\ \hline
\end{tabular}
}
}
\end{table*}

\begin{table*}[t]
\centering
\footnotesize{
\caption{ THE RESULT OF COMPARISONS WITH OTHER DEFENSIVE METHODS(Fashion-mnist ,$L_{\infty}$ = 8/25) }
 \label{table:crossstreetmAPresult5}
\resizebox{400pt}{45pt}{

\begin{tabular}{|c|c|c|c|c|c|c|}
\hline
\multirow{2}{*}{Network}  & \multirow{2}{*}{\begin{tabular}[c]{@{}c@{}}Defensive\\ Method\end{tabular}} & \multirow{2}{*}{Clean} & \multirow{2}{*}{FGSM} & \multirow{2}{*}{BIM} & \multirow{2}{*}{DeepFool} & \multirow{2}{*}{C\&W} \\
                          &                                                                             &                        &                       &                      &                           &                     \\ \hline
\multirow{7}{*}{Resnet50} & Normal                                                                      & 93\%/93\%              & 38\%/24\%             & 00\%/00\%            & 06\%/06\%                 & 00\%/00\%           \\ \cline{2-7}
                          & Adversarial FGSM                                                            & 93\%/93\%              & 85\%/85\%             & 51\%/00\%            & 63\%/07\%                 & 67\%/21\%           \\ \cline{2-7}
                          & Adversarial BIM                                                             & 92\%/91\%              & 84\%/79\%             & 76\%/63\%            & 82\%/72\%                 & 81\%/70\%           \\ \cline{2-7}
                          & Label Smoothing                                                             & 93\%/83\%              & 73\%/45\%             & 16\%/00\%            & 29\%/06\%                 & 33\%/14\%           \\ \cline{2-7}
                          & Feature Squeezing                                                           & 84\%/84\%              & 70\%/28\%             & 56\%/25\%            & 83\%/83\%                 & 83\%/83\%           \\ \cline{2-7}
                          & PiexlDefend                                                                 & 89\%/89\%              & 87\%/82\%             & \textbf{85\%/83\%}           & 88\%/88\%                 & 88\%/88\%           \\ \cline{2-7}
                          & \textbf{Proposed method}                                                             & \textbf{93\%/93\%}              & \textbf{89\%/89\%}             & 70\%/60\%            & \textbf{90\%/89\%}                 & \textbf{88\%/89\%}           \\ \hline
\end{tabular}
}
}
\end{table*}

\begin{table*}[t]
\centering
\footnotesize{
\caption{ THE RESULT OF COMPARISONS WITH OTHER DEFENSIVE METHODS(Imagenet ,$L_{\infty}$ = 8/13/20) }
 \label{table:crossstreetmAPresult6}
\resizebox{400pt}{28pt}{

\begin{tabular}{|c|c|c|c|c|c|c|}
\hline
\multirow{2}{*}{Network}  & \multirow{2}{*}{\begin{tabular}[c]{@{}c@{}}Defensive\\ Method\end{tabular}} & \multirow{2}{*}{Clean} & \multirow{2}{*}{FGSM} & \multirow{2}{*}{BIM} & \multirow{2}{*}{DeepFool} & \multirow{2}{*}{C\&W} \\
                          &                                                                             &                        &                       &                      &                           &                     \\ \hline
\multirow{3}{*}{Resnet101} & Normal                                                                      & 76\%/76\%/76\%         & 03\%/03\%/03\%        & 00\%/00\%/00\%       & 01\%/01\%/01\%            & 00\%/00\%/00\%      \\ \cline{2-7}
                          & HGD                                                                         & 54\%/54\%/54\%         & 51\%/50\%/50\%        & \textbf{36\%/36\%/36\%}       & 52\%/52\%/52\%            & 51\%/51\%/51\%      \\ \cline{2-7}
                          & \textbf{Proposed method}                                                             & \textbf{67\%/67\%/67\%}         & \textbf{56\%/56\%/56\%}        & 12\%/12\%/10\%       & \textbf{53\%/53\%/52\%}            & \textbf{54\%/54\%/53\%}      \\ \hline

\end{tabular}
}
}
\end{table*}

\begin{table*}[t]
\centering
\caption{Comparison results with HGD on ImageNet ($L_{\infty}$ = 8/16)}
 \label{table:crossstreetmAPresult7}
\resizebox{400pt}{56pt}{
\begin{tabular}{|c|c|c|cl|c|c|c|}
\hline
Network                                          & Defense                       & Clean                & FGSM                  & IFGSM(3/5)          &MI-FGSM          & Deepfool                & C\&W                             \\ \hline
\multirow{3}{*}{IncResV2}              & Normal                             & 86\%               & 34\%/30\%             & 10\%/5\%                &   13\%/7\%         & 13\%/11\%             & 0\%/0\%                                \\ \cline{2-8}
                                                       & HGD                                & 54\%               & 47\%/48\%              & 42\%/42\%               &        46\%/44\%          & 48\%/48\%            & 48\%/48\%                             \\ \cline{2-8}
                                                       & Our method            & \textbf{77\%}   & \textbf{62\%/61\%} &\textbf{51\%/42\%}              &         \textbf{50\%}/40\%        & \textbf{60\%/60\%} & \textbf{61\%/63\%}                \\ \hline

\multirow{3}{*}{IncV3}                     & Normal                            & 83\%                & 20\%/18\%             & 57\%/49\%         & 57\%/50\%           &                12\%/11\%             & 0\%/0\%                                  \\  \cline{2-8}
                                                       & HGD                                & 70\%                & 60\%/60\%             & 62\%/\textbf{61\%}                &         62\%/62\%       & 60\%/60\%            & 59\%/59\%                               \\ \cline{2-8}
                                                      & Our method             & \textbf{74\%}   &\textbf {62\%/61\%} & \textbf{64\%}/60\%              &               \textbf{69\%/64\%}      & \textbf{60\%/60\%} & \textbf{60\%/60\%}                   \\\hline

\multirow{3}{*}{IncV4}                   & Normal                               & 88\%               & 28\%/26\%               & 6\%/1\%   &   4\%/1\%            & 17\%/15\%            & 0\%/0\%                                     \\ \cline{2-8}
                                                     & HGD                                   & 64\%               & 56\%/56\%               & \textbf{51\%/50\%}      &   \textbf{57\%/52\%}           & 59\%/59\%              & 59\%/59\%                     \\ \cline{2-8}
                                                     & Our method              & \textbf{74\%}  & \textbf{58\%/56\%}             & 50\%/46\%                   &     50\%/40\%         & \textbf{60\%/60\%} & \textbf{61\%/60\%}                       \\ \hline

\end{tabular}
}
\end{table*}

\subsection{ Selection of hyper parameters}
There are two hyper parameters in the neural networks that
need to be determined by a large number of experiments. The first one is the standard normal distribution gaussian noise parameter
\begin{math}\varphi\end{math}, and the second one is the penalty item parameter \begin{math}\lambda\end{math}. In order to improve the performance of the proposed method, the value of \begin{math}\varphi\end{math} and \begin{math}\lambda\end{math} is depending on the performance
of image classification. Specifically, image compression discards part of the image information even if it retains the main structural information of the image.
In order to keep the accuracy of image classifier, we compute
the average accuracy of  the well-trained Resnet50 on the 1000 random images of the cifar-10 training dataset.  For more details, please refer to Table \ref{table:crossstreetmAPresult3}.
\par From Table \ref{table:crossstreetmAPresult3}, we can see that when the parameter \begin{math}\lambda\end{math}
is fixed, the accuracy of the classifier first increases and then
decreases with the increase of parameter \begin{math}\varphi\end{math}.
More specifically, when parameter  \begin{math}\lambda= 0.0001\end{math} and
\begin{math}\varphi= 1.0 \sim 20.0\end{math}, the average accuracy increases
constantly. But when parameter  \begin{math}\lambda = 0.0001\end{math} and
\begin{math}\varphi= 20.0 \sim 50.0\end{math}, the average accuracy decreases
constantly. That is, when the noise is too large, the network is not enough to resist
it, resulting in a decline in network performance and when the
noise is too small, the network learns to use the gray scale
information between 0 and 1 to encode the image instead of using 0 and 1.
Similarly, when the parameter \begin{math}\varphi\end{math}
is fixed, the accuracy of the classifier first increases and then decreases with the
decrease of parameter \begin{math}\lambda\end{math}. Therefore, the appropriate parameter
settings can protect the accuracy of image classification models.  In
accordance with Table \ref{table:crossstreetmAPresult3}, \begin{math}\lambda= 0.0001\end{math} and
\begin{math}\varphi= 20.0 \end{math} can obtain the best performance of
the classifier. In addition, In this paper, the value of the parameter
\begin{math}\lambda\end{math} is 0.0001 and the value of the parameter
\begin{math}\varphi\end{math} is 20.0.

%-------------------------------------------------------------------------
%-------------------------------------------------------------------------
\subsection{Image classification with the proposed method}
Simply detecting adversarial images is not sufficient for the task of the
image classification. It is often critical to be able to correctly classify
adversarial examples. In this section, there are two scenarios where our proposed method is used to defend the adversarial attacks. One
is using the image compression at test time, the other is using the image compression at training and test time.
%-------------------------------------------------------------------------
\begin{figure}
\begin{center}
   \includegraphics[width=1\linewidth]{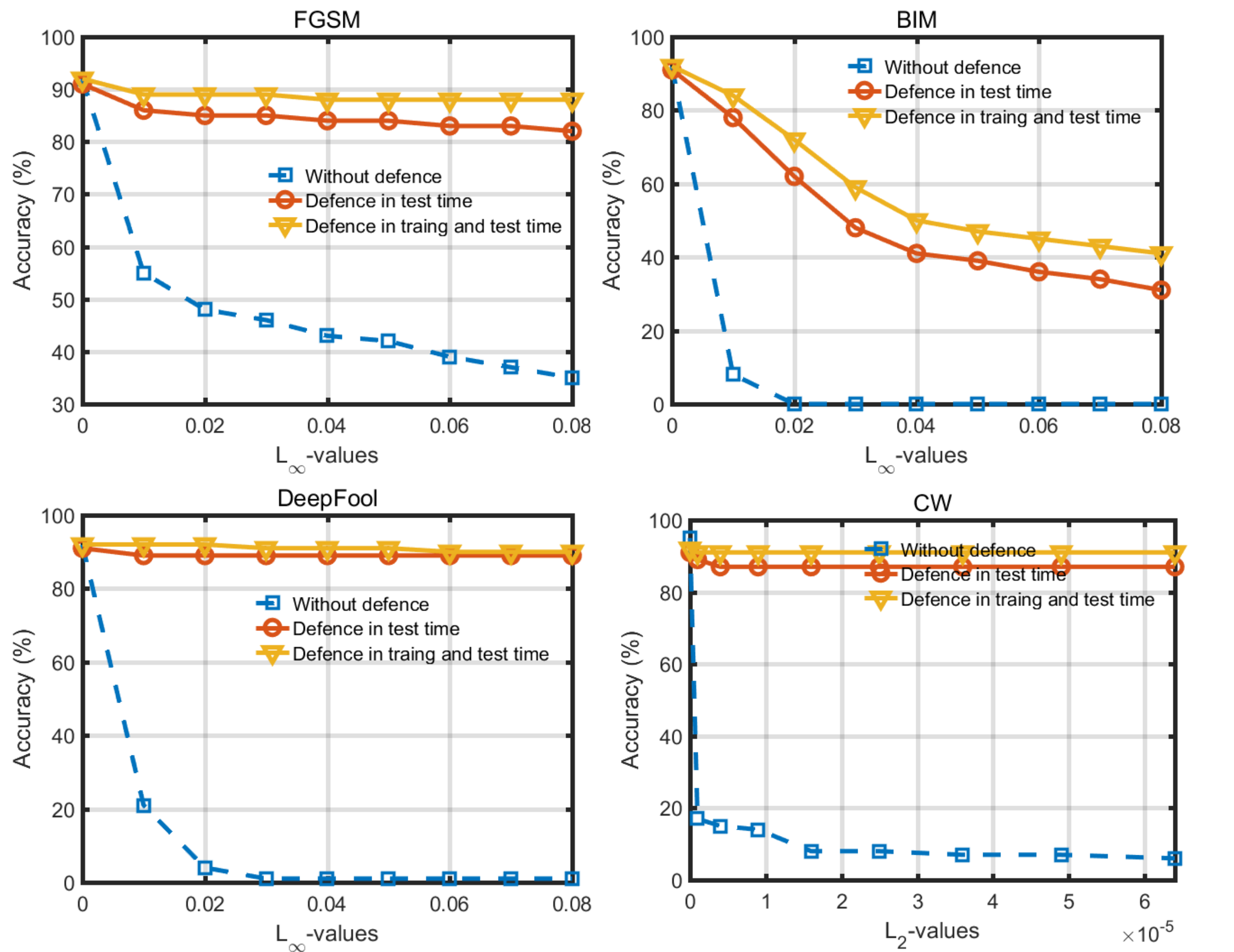}
\end{center}
\setlength{\abovecaptionskip}{-0.2cm}
   \caption{ The classification accuracy of ResNet-50 on adversarial images produced by four
    attacks using the proposed method at the test time and at training and test time. The
    dotted line represents the accuracy of the ResNet-50 model on adversarial images without
    any defense.
    }
\label{fig:jxj}
\end{figure}

%-------------------------------------------------------------------------
\subsubsection{Image compression at test time}
The image classification model has been trained on the clean images. The test images consist of clean images and adversarial images. They are first compressed and reconstructed by the proposed method, and then they are fed to the well-trained classifier. Fig \ref{fig:jxj} shows the accuracy of image classification model(Resnet50) which are tested on the adversarial examples produced by the four attacks. The dotted lines show the accuracy of image classification models tested on the adversarial images with no defense. In this respect, the proposed method using at the
test time increases accuracy on the FGSM strongest attack from 35\% to 83\%, the BIM
strongest attack from 0\% to 31\%, the DeepFool strongest attack from
1\% to 89\% and the C\&W strongest attack from 0\% to 87\% for CIFAR-10.

%-------------------------------------------------------------------------

\begin{table}
\footnotesize{
\caption{ Comparison results with ICLR2018 on ImageNet}
 \label{table:result05}
 \label{table:crossstreetmAPresult9}
\resizebox{240pt}{20pt}{
\begin{tabular}{|c|c|c|c|c|c|}
\hline
Network              & Defensive Method & FGSM-8 & FGSM-12  & Deepfool & C\&W \\ \hline
\multirow{3}{*}{IncResV2} & Normal           & 34\%  & 32\%            & 13\%                          & 0\%                     \\ \cline{2-6}
                          & ICLR2018    & 62\%  & 50\%   & 55\%                   & 59\%                    \\ \cline{2-6}
                          & Our method  & \textbf{62\%}  & \textbf{61\%}       & \textbf{60\%}        & \textbf{61\%}                    \\ \hline
\end{tabular}
}}\vspace{-.5cm}
\end{table}

%-------------------------------------------------------------------------
\subsubsection{Image compression at training and test time}
There is another way to defend the adversarial examples, that is to say, we apply
the proposed method during the training and test time. In particular, During the training time,
we train the image classification models on transformed cifar-10 training images.
We first use the ComCNN to compress the input image into the compact representation,
and then use the RecCNN to reconstruct the input image before feeding it to the network. As for the test time, the test image is transformed by the proposed method before being fed to the well-trained classifier.
As shown in Fig \ref{fig:jxj}, we can see that the proposed method at training and test time
increases accuracy on the FGSM strongest attack from 35\% to 83\%, the BIM
strongest attack from 0\% to 31\%, the DeepFool strongest attack from
1\% to 89\% and the C\&W strongest attack from 0\% to 87\% for CIFAR-10.

%-------------------------------------------------------------------------
\subsubsection{Comparisons with other defensive methods}
In order to quantitatively measure the performance of our proposed
method, we compare the proposed method with other conventional schemes
under the \begin{math} L_{\infty}\end{math} distance metric.
The result of the comparison on the Cifar-10 image dataset is shown in Table \ref{table:crossstreetmAPresult4}. During training and test time, compared
with these methods, our proposed method achieves huge performance improvement.
In particular, it achieves nearly 90\% accuracy on the FGSM, DeepFool and C\&W attack
methods. Compared with the image classification model on the clean images,
the accuracy of the model on the adversarial examples does not decline a lot.
As for defense applied in test time, the proposed method can achieve about 85\% accuracy on the FGSM, DeepFool and C\&W attack methods. As for BIM attack, the performance is improved by using our proposed method. And Table \ref{table:crossstreetmAPresult5} shows the result
of the comparison with other defensive methods on the Fashion-mnist image dataset. The performance
is improved a lot by using the proposed method on FGSM, DeepFool and C\&W attack
methods. More importantly, we do comparison experiment with HGD and ICLR2018\cite{xie2017mitigating} method on the imagenet dataset. As shown in Table \ref{table:crossstreetmAPresult6}, the proposed method improves the performance of defending the FGSM, DeepFool and C\&W attack methods. And to test more attacking methods, we add the deepfool and C\&W methods. The results are shown in Table \ref{table:crossstreetmAPresult7}. We see that the proposed method achieves the higher defensive accuracy against FGSM, DeepFool, MI-FGSM \cite{dong2018boosting} and C\&W, and the competitive accuracy against IFGSM compared with HGD, which demonstrates the effectiveness of the proposed method. Note that the $\epsilon$ for IFGSM is set to 3 and 5, rather 8 and 16, because we find when the attack is too strong (when $L_{\infty}>= 8$), the noises are perceptible to human eyes. And thus, the adversarial examples can be easily distinguished by human beings, and the defense methods are not necessary. \textbf{In fact, the core advantage of our method is that we train our network on the clean images rather than adversarial images. In this way, we don't need to use attacking methods to generate adversarial examples, and thus the training data set is much smaller than HGD, and the training time is also much less than HGD. Besides, our method performes the compression based on the patch rather than the entire image, therefore, the testing time is reduced (the HGD takes 2.7 seconds to process an image. But the proposed method only takes 1.2 seconds to process the same image).} We give the comparison results with \cite{xie2017mitigating} in Table \ref{table:crossstreetmAPresult9}. The results shows that our method can achieve the higher accuracy against deepfool, C\&W, and  FGSM with $\epsilon=8,12$ than \cite{xie2017mitigating}, which verifies the effectiveness of our method. Furthermore, our proposed method does not depend on attacking methods and classifiers. And it can be combined with other defensive methods.

%-------------------------------------------------------------------------
%-------------------------------------------------------------------------
\subsection{Analysis for the proposed method}
For the test time, the proposed method transforms the input image to a clean image.
And it breaks up the particular structure of the perturbations in the adversarial examples.
Specifically, the ComCNN encode the input image into a compact representation. During this process,
the imperceptible perturbations do not affect the result of the compact representation. In other words,
the output images of the clean image and adversarial image are as same as possible.
Because during the training of the network, the network learns to resist the stronger gaussian noise attack to encode the input images. For the training and test time,
the proposed method compresses the space of the real samples. For the uncompressed space,
there are \begin{math}32\times32\times2^{8}\times2^{8}\times2^{8}\end{math} images in this
space. But for the compressed space, there only are \begin{math}32\times32\times2^{4}\times2^{4}\times2^{4}\end{math}
images in this space. In this way, the proposed method makes the existing image classification models easier to simulate the image distribution. The mezzanine is between
the decision surface trained by the classifier and the real surface of the sample data becomes smaller. That is to
say, the probability of the adversarial example occurrence becomes smaller than before.

%-------------------------------------------------------------------------
\section{Conclusion}
In this paper, we propose an end-to-end image compression model to defend adversarial examples.
ComDefend can be used in test time and in training
and test time. As for test time, it defends the adversarial examples by destroying the structure of adversarial perturbations in the adversarial image. As for training and test time, it achieves defense by compressing the image space.
In this way, it reduces the search space available for an adversary to construct adversarial examples. ComDefend can achieve higher accuracy on FGSM, DeepFool and C\&W attack methods compared with the state-of-the-art defense methods. And the performance on BIM attack also improves by using our proposed method. More importantly, ComDefend is performed on an image with the
patch-by-patch manner instead of the whole image, which is taken less time to deal with the input image. Our work demonstrates that the performance of classifying the adversarial examples is dramatically improved by using the proposed method.

%-------------------------------------------------------------------------
\section{Acknowledgements}
Supported by the National Key R\&D Program of China (Grant No.2018YFB0803701). National Natural Science Foundation of China (No.U1636214, 61861166002, U1803264, 61806109). Beijing Natural Science Foundation (No.L182057). Project funded by China Postdoctoral Science Foundation (No.2018M641360).

{\small
\bibliographystyle{ieee}
\bibliography{egbib}

\begin{thebibliography}{10}\itemsep=-1pt

\bibitem{carlini2017towards}
N.~Carlini and D.~Wagner.
\newblock Towards evaluating the robustness of neural networks.
\newblock In {\em 2017 IEEE Symposium on Security and Privacy (SP)}, pages
  39--57. IEEE, 2017.

\bibitem{clevert2015fast}
D.-A. Clevert, T.~Unterthiner, and S.~Hochreiter.
\newblock Fast and accurate deep network learning by exponential linear units
  (elus).
\newblock {\em arXiv preprint arXiv:1511.07289}, 2015.

\bibitem{collobert2008unified}
R.~Collobert and J.~Weston.
\newblock A unified architecture for natural language processing: Deep neural
  networks with multitask learning.
\newblock In {\em Proceedings of the 25th international conference on Machine
  learning}, pages 160--167. ACM, 2008.

\bibitem{deng2009imagenet}
J.~Deng, W.~Dong, R.~Socher, L.-J. Li, K.~Li, and L.~Fei-Fei.
\newblock Imagenet: A large-scale hierarchical image database.
\newblock In {\em Computer Vision and Pattern Recognition, 2009. CVPR 2009.
  IEEE Conference on}, pages 248--255. Ieee, 2009.

\bibitem{dong2018boosting}
Y.~Dong, F.~Liao, T.~Pang, H.~Su, J.~Zhu, X.~Hu, and J.~Li.
\newblock Boosting adversarial attacks with momentum.
\newblock In {\em Proceedings of the IEEE Conference on Computer Vision and
  Pattern Recognition}, pages 9185--9193, 2018.

\bibitem{goodfellow6572explaining}
I.~J. Goodfellow, J.~Shlens, and C.~Szegedy.
\newblock Explaining and harnessing adversarial examples (2014).
\newblock {\em arXiv preprint arXiv:1412.6572}.

\bibitem{he2015delving}
K.~He, X.~Zhang, S.~Ren, and J.~Sun.
\newblock Delving deep into rectifiers: Surpassing human-level performance on
  imagenet classification.
\newblock In {\em Proceedings of the IEEE international conference on computer
  vision}, pages 1026--1034, 2015.

\bibitem{he2016deep}
K.~He, X.~Zhang, S.~Ren, and J.~Sun.
\newblock Deep residual learning for image recognition.
\newblock In {\em Proceedings of the IEEE conference on computer vision and
  pattern recognition}, pages 770--778, 2016.

\bibitem{hinton2012deep}
G.~Hinton, L.~Deng, D.~Yu, G.~E. Dahl, A.-r. Mohamed, N.~Jaitly, A.~Senior,
  V.~Vanhoucke, P.~Nguyen, T.~N. Sainath, et~al.
\newblock Deep neural networks for acoustic modeling in speech recognition: The
  shared views of four research groups.
\newblock {\em IEEE Signal processing magazine}, 29(6):82--97, 2012.

\bibitem{kingma2014adam}
D.~P. Kingma and J.~Ba.
\newblock Adam: A method for stochastic optimization.
\newblock {\em arXiv preprint arXiv:1412.6980}, 2014.

\bibitem{krizhevsky2010cifar}
A.~Krizhevsky, V.~Nair, and G.~Hinton.
\newblock Cifar-10 (canadian institute for advanced research).
\newblock {\em URL http://www. cs. toronto. edu/kriz/cifar. html}, 2010.

\bibitem{kurakin2016adversarial}
A.~Kurakin, I.~Goodfellow, and S.~Bengio.
\newblock Adversarial machine learning at scale.
\newblock {\em arXiv preprint arXiv:1611.01236}, 2016.

\bibitem{kurakin2018adversarial}
A.~Kurakin, I.~Goodfellow, S.~Bengio, Y.~Dong, F.~Liao, M.~Liang, T.~Pang,
  J.~Zhu, X.~Hu, C.~Xie, et~al.
\newblock Adversarial attacks and defences competition.
\newblock {\em arXiv preprint arXiv:1804.00097}, 2018.

\bibitem{lecun2015deep}
Y.~LeCun, Y.~Bengio, and G.~Hinton.
\newblock Deep learning.
\newblock {\em nature}, 521(7553):436, 2015.

\bibitem{liao2017defense}
F.~Liao, M.~Liang, Y.~Dong, T.~Pang, J.~Zhu, and X.~Hu.
\newblock Defense against adversarial attacks using high-level representation
  guided denoiser.
\newblock {\em arXiv preprint arXiv:1712.02976}, 2017.

\bibitem{madry2017towards}
A.~Madry, A.~Makelov, L.~Schmidt, D.~Tsipras, and A.~Vladu.
\newblock Towards deep learning models resistant to adversarial attacks.
\newblock {\em arXiv preprint arXiv:1706.06083}, 2017.

\bibitem{moosavi2016deepfool}
S.-M. Moosavi-Dezfooli, A.~Fawzi, and P.~Frossard.
\newblock Deepfool: a simple and accurate method to fool deep neural networks.
\newblock In {\em Proceedings of the IEEE Conference on Computer Vision and
  Pattern Recognition}, pages 2574--2582, 2016.

\bibitem{song2017pixeldefend}
Y.~Song, T.~Kim, S.~Nowozin, S.~Ermon, and N.~Kushman.
\newblock Pixeldefend: Leveraging generative models to understand and defend
  against adversarial examples.
\newblock {\em arXiv preprint arXiv:1710.10766}, 2017.

\bibitem{szegedy2013intriguing}
C.~Szegedy, W.~Zaremba, I.~Sutskever, J.~Bruna, D.~Erhan, I.~Goodfellow, and
  R.~Fergus.
\newblock Intriguing properties of neural networks.
\newblock {\em arXiv preprint arXiv:1312.6199}, 2013.

\bibitem{tramer2017ensemble}
F.~Tram{\`e}r, A.~Kurakin, N.~Papernot, I.~Goodfellow, D.~Boneh, and
  P.~McDaniel.
\newblock Ensemble adversarial training: Attacks and defenses.
\newblock {\em arXiv preprint arXiv:1705.07204}, 2017.

\bibitem{warde201611}
D.~Warde-Farley and I.~Goodfellow.
\newblock 11 adversarial perturbations of deep neural networks.
\newblock {\em Perturbations, Optimization, and Statistics}, page 311, 2016.

\bibitem{warde2016adversarial}
D.~Warde-Farley, I.~Goodfellow, T.~Hazan, G.~Papandreou, and D.~Tarlow.
\newblock Adversarial perturbations of deep neural networks. perturbations.
\newblock {\em Optimization, and Statistics}, 2, 2016.

\bibitem{xiao2017fashion}
H.~Xiao, K.~Rasul, and R.~Vollgraf.
\newblock Fashion-mnist: a novel image dataset for benchmarking machine
  learning algorithms.
\newblock {\em arXiv preprint arXiv:1708.07747}, 2017.

\bibitem{xie2017mitigating}
C.~Xie, J.~Wang, Z.~Zhang, Z.~Ren, and A.~Yuille.
\newblock Mitigating adversarial effects through randomization.
\newblock In {\em International Conference on Learning Representations}, 2018.

\bibitem{xu2017feature}
W.~Xu, D.~Evans, and Y.~Qi.
\newblock Feature squeezing: Detecting adversarial examples in deep neural
  networks.
\newblock {\em arXiv preprint arXiv:1704.01155}, 2017.

\end{thebibliography}
}

\end{document}